\documentclass[sigconf]{acmart}
\renewcommand\footnotetextcopyrightpermission[1]{} % removes footnote with conference information in first column
\usepackage{booktabs} % For formal tables
\usepackage{algorithm}
\usepackage{algorithmicx} 
\usepackage{algpseudocode}
\usepackage{multirow}
\usepackage{subcaption}
\usepackage{changepage}
\usepackage{diagbox}
\setcopyright{none}
\begin{document}
\title{Face Recognition with Hybrid Efficient Convolution Algorithms on FPGAs}
%\titlenote{Produces the permission block, and copyright information}
%\subtitle{Extended Abstract}
%\subtitlenote{The full version of the author's guide is available as   \texttt{acmart.pdf} document}

\author{Chuanhao Zhuge$^{1}$, Xinheng Liu$^{1, 3}$, Xiaofan Zhang$^{1}$,     		  Sudeep Gummadi$^{1}$, \\ 
		Jinjun Xiong$^{2}$, Deming Chen$^{1, 3}$\\% <-this % stops a space 
        $^{1}$University of Illinois Urbana-Champaign\\
        $^{2}$T. J. Watson Research Center, IBM,
        $^{3}$Inspirit IoT, Inc.
}

\begin{abstract}
\quad Deep Convolutional Neural Networks (CNN) have become a Swiss knife in solving critical artificial intelligence tasks. However, deploying deep CNN models for latency-critical tasks remains to be challenging because of the complex nature of CNNs. Recently, FPGA has become a favorable device to accelerate deep CNNs thanks to its high parallel processing capability and energy efficiency. In this work, we explore different fast convolution algorithms including Winograd and Fast Fourier Transform (FFT), and find an optimal strategy to apply them together on different types of convolutions. We also propose an optimization scheme to exploit parallelism on novel CNN architectures such as Inception modules in GoogLeNet. We implement a configurable IP-based face recognition acceleration system based on FaceNet using High-Level Synthesis. Our implementation on a Xilinx Ultrascale device achieves 3.75x latency speedup compared to a high-end NVIDIA GPU and surpasses previous FPGA results significantly.
\end{abstract}




\maketitle

\section{Introduction}
\quad Deep Convolutional Neural Networks (CNN) have demonstrated great success in various machine intelligence areas and enabled new advancement for applications such as video content understanding \cite{donahue2015long}, face recognition \cite{schroff2015facenet}, and crowd flow monitering \cite{li2018csrnet}. To overcome the overwhelming computing pressure of these deep models, researchers have developed custom hardware accelerators including ASICs \cite{chen2014diannao} and FPGAs \cite{qiu2016going}. FPGAs have already been proven to be an efficient device for implementing traditional computer vision algorithms \cite{liu2011real, rupnow2011high, he2009novel}. More recently, they also gained popularity in deep neural network accelerations mainly because of their flexibility and high energy-efficiency. And with the help of High-Level Synthesis (HLS) tools, we are able to rapidly map and optimize the emerging deep neural network architectures on FPGAs.\\
\indent Prior FPGA works \cite{zhang2015optimizing, ovtcharov2015accelerating} mostly focus on optimizing general convolution (CONV) layers. However, recent trend shows that state-of-the-art neural network architectures \cite{szegedy2015going, abdi2016multi} tend to contain the topology of parallel branches, which are then merged through filter concatenation or summation. It is noticed that jointly optimizing a single acceleration engine for all layers of the deep networks leads to dynamic underutilization of resources \cite{shen2016overcoming}. This effect is especially acute on these advanced CNN structures with branches, since each branch would most likely carry different sizes and dimensions of convolutions. Also, past approaches did not exploit the attribute that individual branches are independent, thus do not utilize the opportunity to concurrently execute the parallelizeable CONV branches. To overcome this problem, one solution is to implement multiple CONV engines that are specifically designed for each or subsets of layers. To achieve minimum overall latency, a resource partition solution has been proposed \cite{zhang2017high}. We extend such resource allocation strategy to exploit the intra-module parallelism of the multi-branch topology to attain better latency.\\
\indent Besides hardware-specific tricks, researchers also look into algorithmic improvements to accelerate convolution computation. Researchers \cite{vasilache2014fast} exploit the equivalency between convolutions in spatial domain and element-wise multiplications in frequency domain. This method allows us to mathematically reduce the computation complexity with FFT-based convolution. More recently, the Winograd minimal-filter based convolution algorithm has been introduced \cite{lavin2016fast}, and it is suitable for small kernel sizes and strides. Although works have been done to adopt the FFT-based \cite{zhang2017frequency} and Winograd-based \cite{aydonat2017opencl} algorithms to accelerate convolution on FPGAs, to the best of our knowledge, there is no work on combining the two fast algorithms to adapt to different sizes of convolutions so as to obtain better performance. In our work, we analyze and explore the properties of both FFT and Winograd-based convolution algorithms, and propose a heuristic methodology to design a hybrid accelerator for different convolutions. To summarize, this work highlights the following contributions:\\ 
\begin{itemize}
\vspace{-0.1in}
\item We analyze the FFT and Winograd convolution algorithms, and explore the design space to find the suitable sizes and depths for applying each algorithm respectively. The analysis is incorporated in a general methodology to design a hybrid convolution algorithm for FPGAs.
\item We propose a novel resource allocation scheme that considers the intra-module parallelism of the recently invented CNN topology with branches, and minimize the overall system latency. We implement an algorithm to quickly find optimal resource partition parameters on HLS tools.
\item We design a template based reconfigurable HLS IP specifically targeting the Inception module, which features parallel convolution branches \cite{szegedy2015going}. Using the IP, we implement a face recognition system that is built upon Inception V2. We achieve better performance and energy-efficiency compared to GPUs and previous implementation of GoogLeNet on FPGAs \cite{zhang2017frequency}. 
\end{itemize}
\quad The rest of the work is organized as follows. In Section \ref{sec2}, we introduce the background of Inception architecture and two fast convolution algorithms. In Section \ref{sec_dse}, we perform design space exploration on the two algorithms to identify sweet spots to adopt each algorithm. Section \ref{sec_resource} presents the resource allocation scheme along with an algorithm for fast optimization method. The overall implementation on the FPGA and additional techniques for optimization, along with result evaluation, are shown in Section \ref{sec_result}.

\section{Background}\label{sec2}
\subsection{Inception Module}\label{sec2.1}
\begin{figure}[h]
\centering
\includegraphics[width=0.4\textwidth]{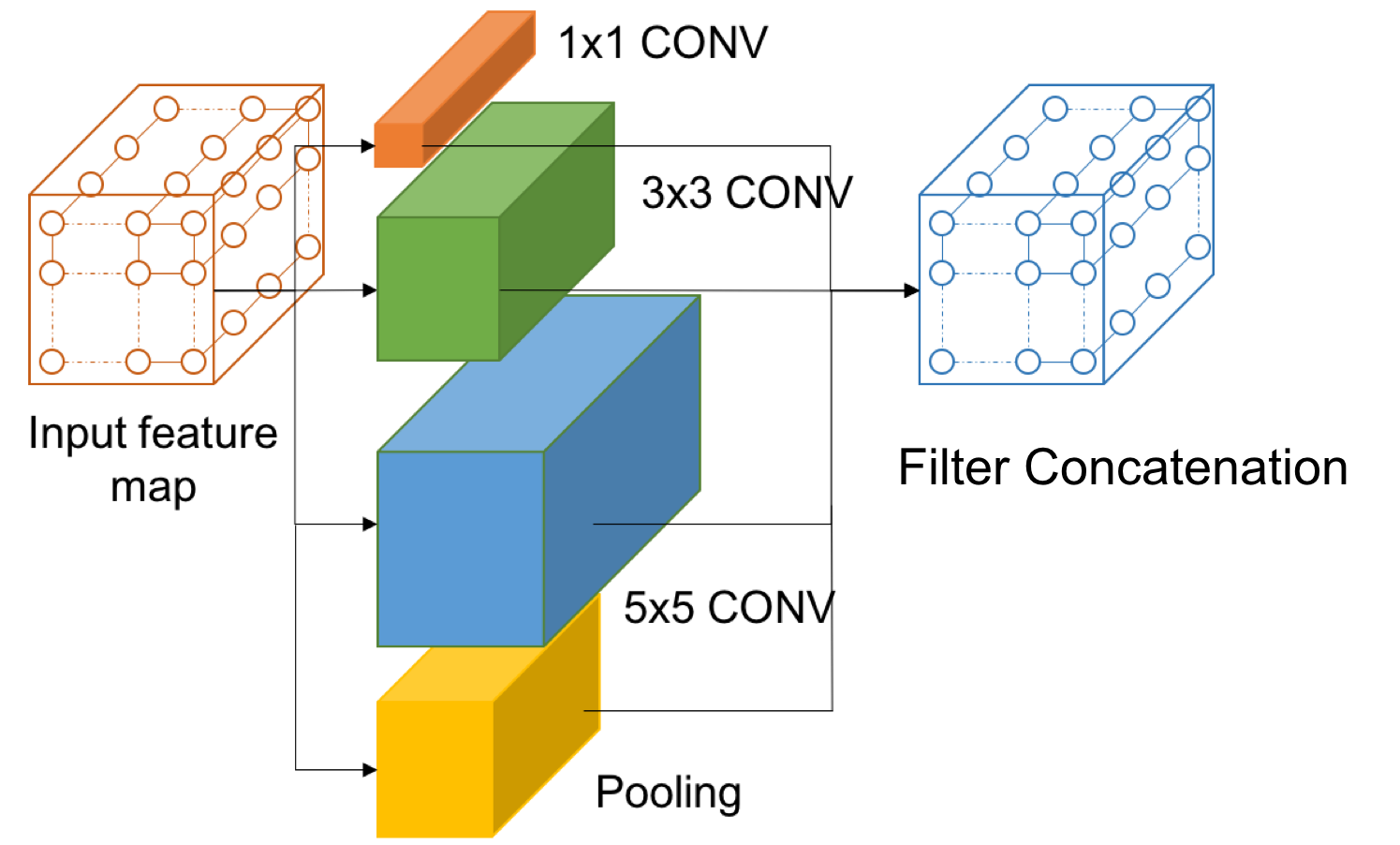}
\caption{Inception module topology.}
\label{fig_inception}
\end{figure}

\begin{figure}[h]
\centering
\includegraphics[width=0.52\textwidth]{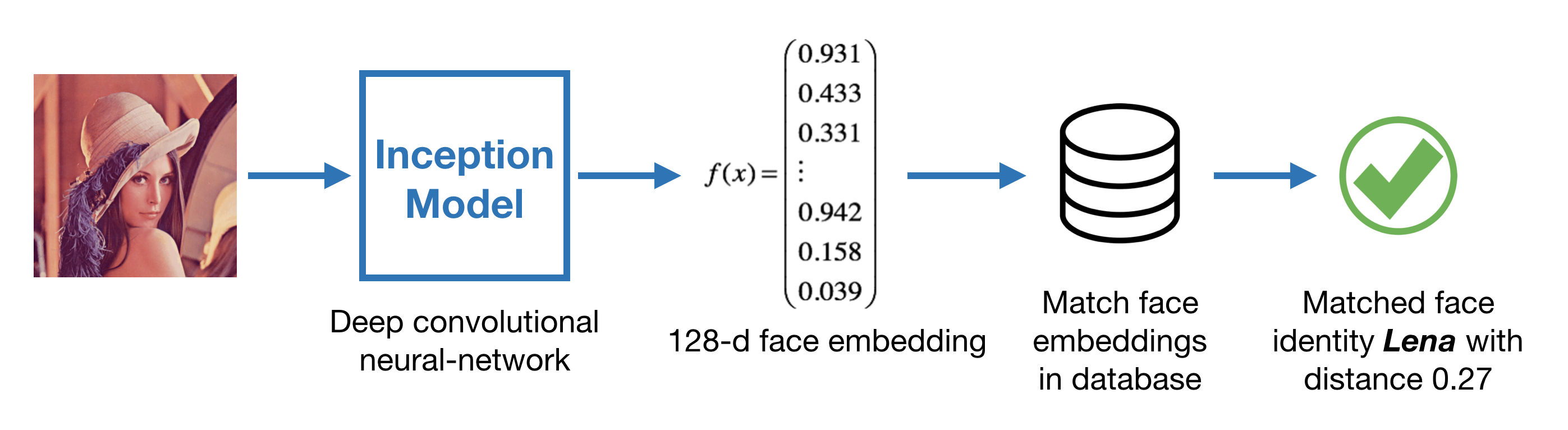}
\caption{Face recognition pipeline.}
\label{fig_fn}
\vspace{-0.1in}
\end{figure}

\quad In recent years, we have seen the booming of  highly effective CNN architectures. One trend is that many networks apply the idea of splitting convolution layers into several branches, which may contain different sizes and depths of convolution kernels. These branches are often merged through concatenation or summation. Such a topology enhances the model's expressivity and enables the network to be several times deeper. The Inception module is the first architecture that employs such a forking mechanism. One inception module contains a composition of pooling, $1 \times 1$, $3 \times 3$, $5 \times 5$ convolutions. At the top, the results of different convolution branches are concatenated together. The Inception module is illustrated in Figure \ref{fig_inception}. The GoogLeNet (Inception V1) comprises nine Inception modules, making it the best performing CNN architecture in ImageNet competition 2014.\\
\quad The FaceNet face recognition system \cite{schroff2015facenet} is based on Inception V2, which is an improvement of V1 with batch normalization. The model is designed to output a 128-d embedding vector, and a typical face recognition pipeline is shown in Figure \ref{fig_fn}. The model is trained using the triplet loss function, such that embeddings of two images of the same person have a small distance between them, while embeddings of different persons have a large distance.

\subsection{Convolution in Frequency Domain}\label{sec2.2}
\quad The well known convolution theorem states that spatial convolutions are equivalent to pair-wise multiplications in the frequency domain. Assuming the convolution input is a $L \times M \times M$ feature map, there are $K$ kernels with size $L \times Q \times Q$. The spatial convolution's computation complexity is then $C_{spatial}=(KLQ^2M^2)$. 2D real-number FFT has logaritmic complexity. During inference, weights are usually only loaded once, therefore the FFT for weights (CONV kernels) can be done offline. Thus, the computation complexity for FFT-based convolution during inference consists of three parts: 2D FFT for feature map, pair-wise complex number multiplication, and inverse-FFT for the result. The overall computation complexity is given by:\\
\begin{center}
\vspace*{-0.25in}
\begin{equation}
\label{fft-conv-complex}
C_{fft\_conv\_inference} = LM^2logM^2 + KL4M^2 + KM^2logM^2 
\end{equation}
\end{center}
And the theoretical speed up is presented as:\\
\begin{center}
\vspace*{-0.25in}
\begin{equation}
\label{fft-speedup}
SU = \frac{KLQ^2}{(K+L)logM^2 + 4KL} \approx \frac{1}{4}Q^2
\end{equation}
\end{center}
when number of feature maps (L) and number of kernels (K) are both large. The feature map size (M) is on log scale thus the impact is minimal. According to Equation \ref{fft-speedup}, larger kernel size leads to more significant speed up.

\subsection{Winograd Minimal Filtering Convolution}\label{sec2.3}
\quad Another fast convolution is based on the Winograd minimal filtering algorithm\cite{winograd1980arithmetic}. The algorithm reduces the number of multiplications with the expense of additional addition and constant multiplication. Take $F(2,3)$ as an example. The standard convolution consumes $2 \times 3 = 6$ multiplications. The Winograd algorithm uses 4 multiplications. It uses 9 more constant multiplications, but they can be implemented as bit-shifts and additions, thus are much cheaper. The 2D Winograd algorithm is implemented from nesting the minimal 1D algorithm. In general, a 2D Winograd algorithm $F(m \times m, r \times r)$ can be represented by the following equations.\\
\begin{center}
\vspace*{-0.2in}
\begin{equation}
U = GgG^T, V = B^TdB, Y = F(m \times m, r \times r) = A^T(U \odot V)A,
\end{equation}
\end{center}
where g, d refer to the original weight tile and feature map tile respectively, G, B, and A are transform matrices, generated by Cook-Toom algorithm, and U, V are the transformed weight tile and feature map tile. For example, $F(2 \times 2, 3 \times 3)$ consumes 16 multiplications, at the expense of additional 84 operations, yielding 2.25x reduction compared to 36 multiplications with standard convolution.
\begin{figure}[h]
\centering
\includegraphics[width=0.35\textwidth]{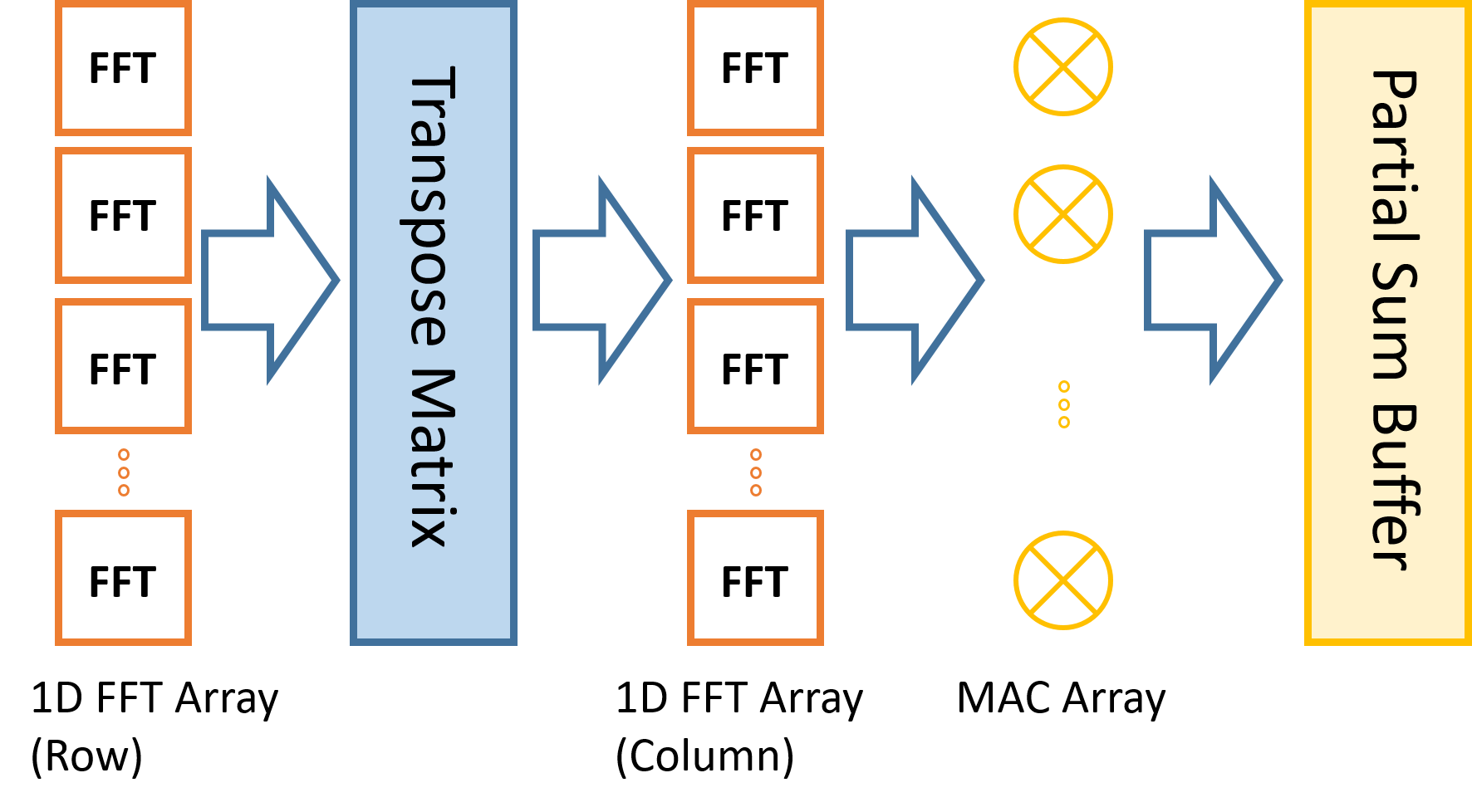}
\caption{2D FFT engine constructed from multiple 1D FFT cores.}
\label{fig_2dfft}
\end{figure}
\begin{figure}[h]
\centering
\includegraphics[width=0.4\textwidth]{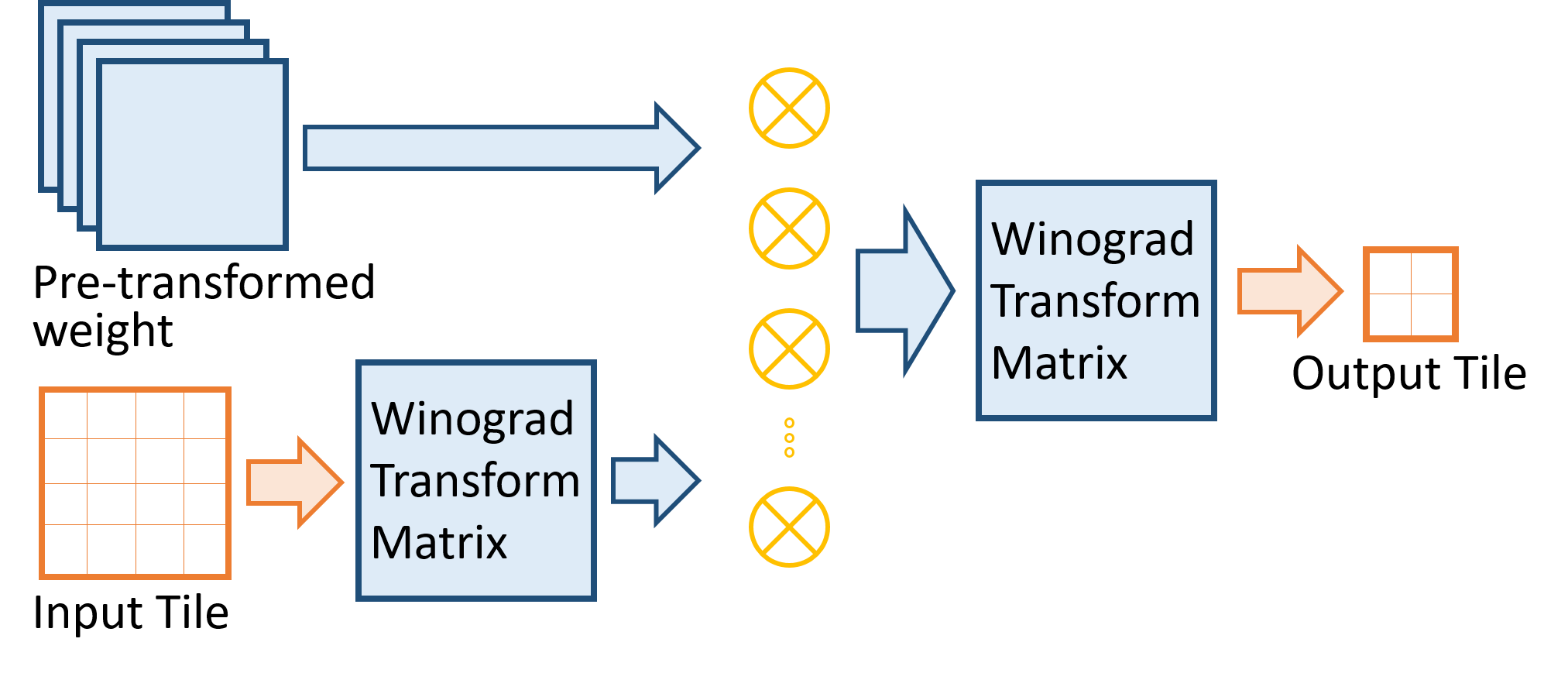}
\caption{Winograd processing engine that exploits parallelism and data reuse.}
\label{fig-winoengine}
\end{figure}

\section{Systemic Characterization for Fast Convolution Algorithms}\label{sec_dse}
\begin{figure}[b]
\centering
\includegraphics[width=0.5\textwidth]{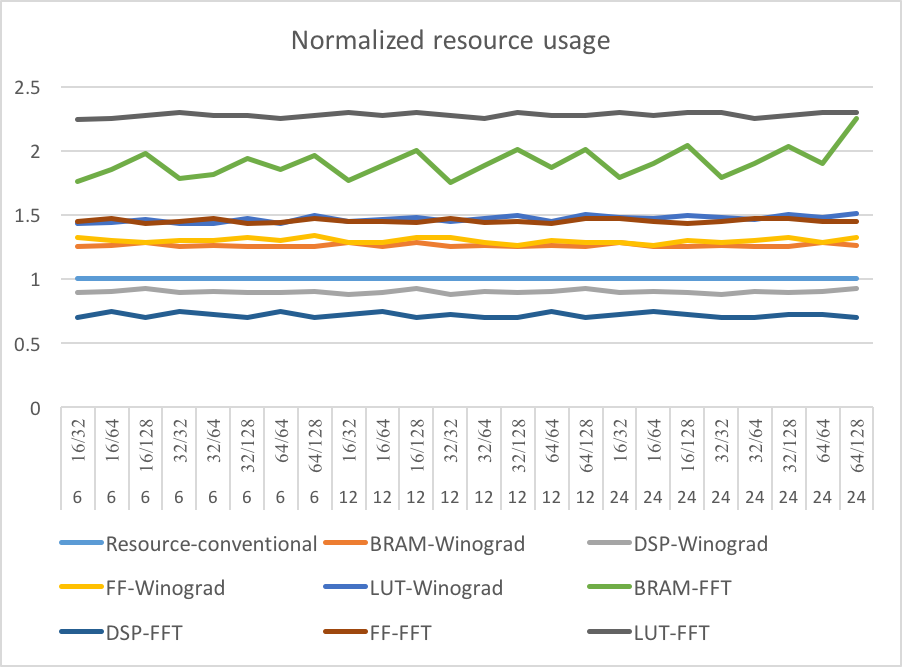}
\caption{Normalized resource usage for different experiment setup.}
\label{fig_rs}
\vspace*{-0.1in}
\end{figure}

\quad The two fast algorithms deliver remarkable speed compared to conventional convolution, but theoretical analysis and previous experiments \cite{vasilache2014fast, lavin2016fast} have shown that these two algorithms have different optimal design points. FFT-based method in theory provides greater speed up when kernel size is larger. On the other hand, study claims that the Winograd algorithm's improvement on speed winds down quickly when kernel size becomes large because the transformation overhead increases quadratically, offsetting the savings in the multiplications.\\
\indent Recent deep CNN structures contain multiple parallel branches with different kinds of convolutions. Therefore, a single efficient algorithm cannot provide the best optimization. Consequently, we come up with an innovative heuristic to design a hybrid accelerator that incorporates both fast algorithms to cover different workloads for better performance. In order to find a good strategy of using different algorithms, we systematically conduct studies for different implementations of FFT and Winograd CONV with different design parameters, and use latency cycle count as our performance metric. In our experiments we consider kernel sizes, feature map sizes, and input/output dimensions shown in Table \ref{dse_config}. The empirical latency model obtained from the study can be used as a guidance to choose different algorithms for different network architectures.\\
\begin{table}[b]
\caption{Design parameters for FFT-based and Winograd-based convolutions}
\centering
\begin{tabular}{ccc}\label{dse_config}
\bf{Dimensions} & \bf{Sizes evaluated}  \\
\hline
kernel sizes & 3, 5, 7 \\
feature map sizes & 6, 12, 24 \\
input/output dimensions & 16, 32, 64, 128 (combinations)
\end{tabular}
\vspace*{-0.2in}
\end{table}
\indent For Winograd-based convolution with larger kernels, we evaluate $F(2 \times 2, 5 \times 5)$, and $F(2 \times 2, 7 \times 7)$. For our implementation of FFT-based convolution, since the Radix-2 FFT inputs must be of size of powers of 2, we choose the padding to make up size 8, 16, and 32, for input size 6, 12, and 24 respectively.\\
\indent One observation is that the kernel size does not affect FFT's absolute performance in general because kernel and input need to be zero-padded to be the same size. There is one exception when input size is $6 \times 6$ and kernel size is $5 \times 5$ and $7 \times 7$. For these particular parameter combinations, we pad it to $16 \time 16$ instead of $8 \times 8$, to retain higher numerical precision as observed in our experiment. The padding overhead is significant, leading to similar performance as $12 \times 12$ input, thus less speed up compared to other algorithms.\\
\indent We use Vivado HLS to implement the algorithms on the Xilinx VU9P FPGA. The implementations of FFT and Winograd CONV engines are illustrated in Figure \ref{fig_2dfft} and \ref{fig-winoengine}. For each engine, they have respective transformation matrices, followed by MAC arrays for computing pair-wise multiplication. Our goal here is to evaluate the algorithmic impact, so we aim to eliminate the hardware resource usage difference as much as we can. We notice that both fast algorithms exhibit similar transform-compute-transform computation pattern, and each transformed feature map can be reused to do multiple pair-wise multiplication with pre-transformed kernels, to generate multiple partial results in parallel. Thus, one easier way to control resource usage is to designate the feature map reuse factor through \textit{HLS UNROLL}. We try to keep the same reuse factor so that different algorithms uses similar amount of resource for fair comparison. However, as shown in Figure \ref{fig_rs}, which demonstrates the normalized resource usage, we observe that different algorithms have different preferences of resources. For example, for FFT-based algorithm, it uses 60\% of the DSPs, but consumes as much as 2.2x of LUTs, compared to the baseline, which is implemented using a conventional loop-optimization method \cite{zhang2017high}. Also, BRAM usage is affected by the number of output channels, since it buffers the intermediate results to prevent unnecessary IFFTs. In general, the fast algorithms prefer using LUTs to implement transformation operations, and save DSP usage due to reduced number of multiplications.\\
\begin{figure}[h]
\centering
\includegraphics[width=0.5\textwidth]{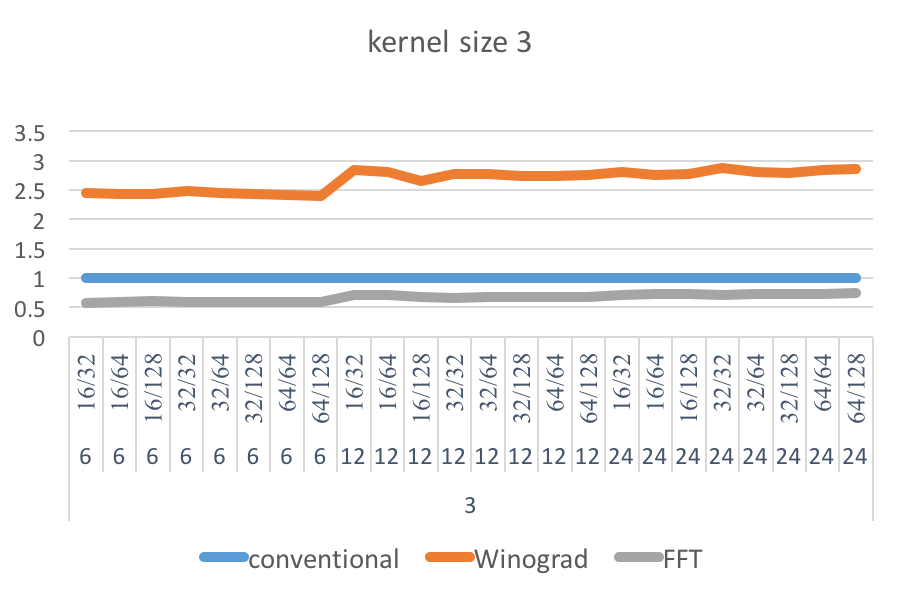}
\vspace*{-0.2in}
\caption{Speed up comparison when kernel size is $3 \times 3$.}
\label{fig_k3}
\vspace*{-0.1in}
\end{figure}
\begin{figure}[t]
\centering
\includegraphics[width=0.5\textwidth]{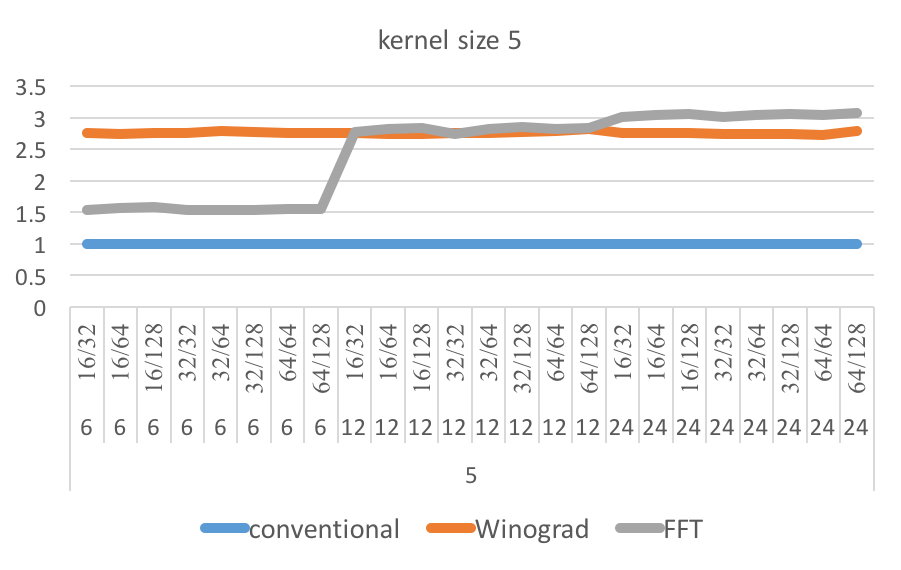}
\vspace*{-0.2in}
\caption{Speed up comparison when kernel size is $5 \times 5$.}
\label{fig_k5}
\vspace*{-0.2in}
\end{figure}
\begin{figure}[h]
\centering
\includegraphics[width=0.5\textwidth]{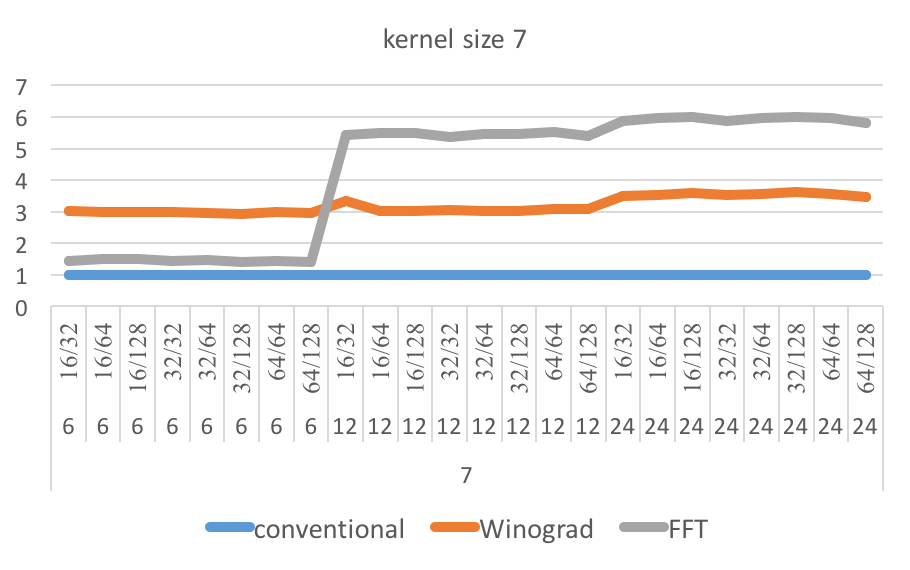}
\vspace*{-0.2in}
\caption{Speed up comparison when kernel size is $7 \times 7$.}
\label{fig_k7}
\vspace*{-0.1in}
\end{figure}
\indent The result is shown in Figure \ref{fig_k3}, \ref{fig_k5}, and \ref{fig_k7}. The X-axis shows the input/output channel sizes, and 2D feature map sizes (6 denotes $6 \times 6$, etc.) of the convolution, and the Y-axis shows the normalized performance of each convolution method, measured in terms of simulation cycle count. Across the three figures, orange curves represent Winograd-based convolution's speed up against the baseline, and grey curves represent FFT-based method's speed up compared to the baseline. From the figure we learn that in small kernels, Winograd's algorithm dominates the performance. For larger kernel sizes, FFT-based convolution starts to catch up in speed, because the expense of Winograd transformation starts to overwhelm. Starting from kernel size $5 \times 5$ and feature map size $24 \times 24$, FFT starts to gain advantage over Winograd. When kernel size is $7 \times 7$ and input/output depth is large, FFT method outperforms Winograd's method by a maximum of 2x margin. Considering that FFT generally has higher resource usage overhead with same unroll factor, it would be wise to apply FFT when kernel size is at least $5 \times 5$ and input size larger than 12. We summarize the empirical results at Table \ref{decisiontable}, which serves to be a decision table to select fast algorithms depending on different input configurations.
\begin{table}[h]
\caption{Decision table for algorithm selection}\label{decisiontable}
  \centering
  \begin{tabular}{|c|c|c|c|}
    \hline
    \diagbox[width=2.2cm, height=2cm]{kernel size}{feature map\\ size} & $6\times6$ & $12\times12$  & $24\times24$ \\
    \hline
    $3\times3$ & Winograd &  Winograd &  Winograd \\
    \hline
    $5\times5$ & Winograd & Winograd & FFT \\
    \hline
    $7\times7$ & Winograd  & FFT & FFT \\
    \hline
  \end{tabular}
\vspace*{-0.1in}
\end{table}
\section{Resource Allocation for Minimal Latency Considering Intra-module Parallelism}\label{sec_resource}
\quad With the empirical model, we further develop a judicious resource partition algorithm, which is critical to achieve minimal latency for mapping the entire network onto FPGAs. Zhang's work \cite{zhang2017high} uses Cauchy inequations to prove that in order to optimize the overall latency for an entire network, we should partition the resources according to the equation $\frac{R_i}{R_j} = \frac{\sqrt{C_i}}{\sqrt{C_j}}$, where $\sqrt{C_i}$ and $\sqrt{C_j}$ represent the computation complexity of layer i and j, and $R_i$ and $R_j$ is the calculated ideal resource allocation for layer i and j. Applying such equation with the constraint $R_{total}=\sum_i{R_i}$, one is able to find the optimal resource partition between layers to achieve minimum latency.\\
\indent However, such framework does not consider the properties of Inception like CNN structures with parallel branches, and does not provide us a solution on intra-module resource allocation. In such structure, latency is constrained by the longest branch. We appreciate the fact that each branch in such topology has no dependency and thus executes concurrently. In order to minimize the latency, we prorate the resource according to computation complexities of each branch: $\frac{R_{Bi}}{R_{Bj}} = \frac{C_{Bi}}{C_{Bj}}$, where $Bi$ and $Bj$ indicate different branches.\\
\indent To quickly find out the most appropriate parallel factors for branched structures, we put forward a resource allocation scheme explorer, as shown in Algorithm \ref{algo_exp}. We first calculate the ideal allocation solution for each module as $R_{ideal\_i}$ by solving $\frac{R_i}{R_j} = \frac{\sqrt{C_i}}{\sqrt{C_j}}$ for each layer in the whole network model, then for each layer, we solve $\frac{R_{Bi}}{R_{Bj}} = \frac{C_{Bi}}{C_{Bj}}$ for each branch. In FPGA implementations, to utilize resource more efficiently, a common way is to specify parallel factors to launch multiple computing engines (Winograd or FFT-conv engines in our case) concurrently. To avoid on-chip memory port contention, arrays must be partitioned proportionally to the parallel factors. We designate our parallel factor to follow the power of two, such as 4, 8, and 16. This intends to boost the computing efficiency in hardware and avoid the misaligned parallelism between neighbouring layers (in HLS implementation, array partition factors should be consistent between adjacent layers). The proposed algorithm generates resource allocation scheme for different branch $i$ to approach the theoretical optimum. The algorithm is depicted in Algorithm \ref{algo_exp}. By analyzing the computation demands of the branches, we have the normalized computation complexity (line 1 to 3) and then generate the ideal resource allocation scheme for branch $i$: $R_{ideal\_Bi}$. Since hardware implementation is more favorable to the power of two, we truncate the ideal scheme to use a more realistic option: $R_{Bi}$. From line 7 to 17, we use a while loop to fine-tune the resource allocated to each branch. If the gap between ideal and realistic resource allocation for a branch is non-zero and more resource is still available, we double the current resource utilization (line 10, 11) for the branch to remove this gap starting with the largest gap first to fulfill the computation demand for this critical branch.  

\begin{algorithm}[h]
\footnotesize
\large
\caption{Resource Allocation for Intra-layer Branches}
\label{algo_exp}
  \begin{algorithmic}[1]      
  	\State Calculate computation complexity of each concurrent branch: $C_{B1}, C_{B2}, ..., C_{Bn}$
    \State Normalize the computation complexity: 
    \Statex $C_{norm\_Bi}=\frac{C_{Bi}}{min(C_{B1}, C_{B2}, ..., C_{Bn})}$
    \State Sum up normalized computation complexity: 
    \Statex $C_{norm}=\sum_i\frac{C_{Bi}}{min(C_{B1}, C_{B2}, ..., C_{Bn})}$
    \State Calculate the ideal resource allocation for branch i: \Statex $R_{ideal\_Bi}=R_{total}*\frac{C_{norm\_Bi}}{C_{norm}}$
    \State \textbf{set} $R_{ideal\_sum}=\sum{R_{ideal\_Bi}}$
    \State \textbf{set} $R_{Bi}=2^{\lfloor\log2(R_{ideal\_Bi})\rfloor}, R_{sum}=\sum{R_{Bi}}$
    \State \textbf{While} $R_{sum} < R_{ideal\_sum}$:
    \State Form queue Q based on the utilization gap $D_{Bi}$:
	\Statex \ \ \ \ $D_{Bi}=R_{ideal\_Bi}-R_{i}$ 
	\State Find the largest gap and allocate more resource
    \Statex \ \ \ \ $D_{max}=max(D_{Bi})$, $sel = i$
    \State \ \ \ \ \ \textbf{if} $R_{sum} + R_{sel} <= R_{ideal\_sum}$, 
    \State \ \ \ \ \ \ \ \ $R_{sel} *= 2$, Continue
    \State \ \ \ \ \ \textbf{Endif}
    \State \ \ \ \ \ Delete branch sel in Q;
    \State \ \ \ \ \ \textbf{if} Length(Q) = 0
    \State \ \ \ \ \ \ \ \  Break
    \State \ \ \ \ \ \textbf{Endif}
    \State \textbf{Endwhile}
  \end{algorithmic}
\end{algorithm}
\section{Inception Module IP}\label{incep-ip}
\quad We use Vivado HLS to implement a C++ template based reconfigurable Inception module IP, which includes all the techniques of optimizations discussed in the above sections. Using the Inception engine, a face recognition network, based on Inception V2, is mapped onto a Xilinx VU9P FPGA development board. The Inception module IP is implemented as a template function, with the $cccp$ (1 $\times$ 1 CONV), $conv3$, $conv5$, and $pool$ sub-functions that represent the parallel branches in the Inception module. Although there are no data read/write dependencies between each sub-function, by default, Vivado HLS won't schedule sub-functions to execute concurrently if sub-functions read from or write to the same array (even when they are writing to different location of the array). To solve this problem, we have to explicitly implement $split\_input$ and $combine\_output$ functions to copy the input feature maps to different buffers, and write to isolated buffers for different sub-functions, and then concatenate the result at the end. Algorithm \ref{algo_incep} describes the implementation,
\begin{algorithm}[h]
\footnotesize
\large
\caption{Inception Module IP}
\label{algo_incep}
  \begin{algorithmic}[1]
    \State template $<$typename T0, config\_t CONFIG$>$ 
    \State void inception(T bottom, T top, T weights, T bias) \{
    \State \ \ \ \ T bottom\_conv3, bottom\_conv5, bottom\_cccp;
    \State \ \ \ \ T top\_conv3, top\_conv5, top\_cccp; // buffers
    \State \ \ \ \ \textbf{split\_input}$<$CONFIG$>$(bottom, bottom\_conv3, ...);
    \State \ \ \ \ \textbf{if}(CONFIG.CONV3==1) conv3$<$CONFIG$>$($...$);
    \State \ \ \ \ \textbf{if}(CONFIG.CONV5==1) conv5$<$CONFIG$>$($...$);
    \State \ \ \ \ \textbf{if}(CONFIG.POOL==1) pool$<$CONFIG$>$($...$);
    \State \ \ \ \ \textbf{if}(CONFIG.CCCP==1) cccp$<$CONFIG$<$($...$);
    \State \ \ \ \ \textbf{combine\_output}$<$CONFIG$>$(top, conv3\_top, $...$);
    \State \}
  \end{algorithmic}
\end{algorithm}
where \textit{CONFIG} stands for an ensemble of multiple reconfigurable variables such as array sizes, flags for existence of sub-modules, and unroll factors etc. When Vivado HLS reads the code, it instantiates sub-modules according to the flags in the \textbf{if} statements. With the Inception module IP, we instantiate Inception modules that fit different input/output and CONV sizes by passing template parameters and construct the entire FaceNet system.

\section{System Implementation and Evaluation}\label{sec_result}
\subsection{Data Quantization and Numerical accuracy}
\quad In recent years researchers have shown that neural networks are exceptionally robust to low precision computation \cite{gysel2016hardware}. In this work, we explore both 16-bit fixed point and 8-bit fixed point. We measure the numerical error of the network with regular convolutions, and fast algorithms, whose implementation configuration is listed on Table \ref{detail}. We set the float-point embedding as the ground truth, and measure $l_2$ squared distance between the output with quantized data and float-point data, since this metric is used in the FaceNet system to measure if faces are from the same identity or not. Results are shown on Table \ref{numerical}. First we discover that 8-bit fixed-point weights are adequate to not lose too much accuracy compared with floating-point results. The network with fast algorithms and fixed 16-bit and fixed 8-bit values generate distance error at ~$10^{-4}$ and ~$10^{-1}$ magnitude compared to the floating-point version, which is tolerable in terms of face verification accuracy (threshold for same identity is set to 1). In our experiment of 100 face pairs at size $112 \times 112$, 12 pairs are classified differently compared to the float-point result with 8-bit fast algorithms, and only one mismatched pair with 16-bit fast algorithm. We believe that if we were to run re-training, accuracy can be further restored, as shown in previous works \cite{han2015deep, zhang2017machine}. 
\begin{table}[h]
\caption{Numerical analysis on Inception 3a}\label{numerical}
  \centering
  \begin{tabular}{|p{6.5em}|p{6.5em}|p{6.5em}|p{6em}|}
    \hline
    \textbf{FIX16 L2 error regular CONV} & \textbf{FIX8 L2 error regular CONV}  & \textbf{FIX16 L2 error fast CONV} & \textbf{FIX8 L2 error fast CONV} \\
    \hline
    7.024e-5 & 9.989e-2 & 1.232e-4 & 2.031e-1 \\
    \hline
  \end{tabular}
\vspace*{-0.1in}
\end{table}
\subsection{Implementation}
\quad We implement our design targeting a Xilinx VU9P FPGA. The detailed implementation scheme of our Inception modules with hybrid algorithms is presented in Table \ref{detail}, where "-" means the sub-branch does not exist in the module.\\
\begin{table}[b]
\caption{Inception V2 (GoogLeNet-BN) detailed implementation}\label{detail}
  \centering
  \begin{tabular}{|p{6em}|c|c|}
    \hline
    \textbf{Inception \#} & {$3 \times 3$}  & {$5 \times 5$} \\
    \hline
    \textbf{Inception 2} & {Winograd $F(4\times4, 3\times3)$}  & {-} \\
    \hline
    \textbf{Inception 3a} & {Winograd $F(4\times4, 3\times3)$}  & {conventional} \\
    \hline
    \textbf{Inception 3b} & {Winograd $F(4\times4, 3\times3)$}  & {FFT-based} \\
    \hline
    \textbf{Inception 3c} & {conventional}  & {conventional} \\
    \hline
    \textbf{Inception 4a} & {Winograd $F(2\times2, 3\times3)$}  & {FFT-based} \\
    \hline
    \textbf{Inception 4e}  & {conventional} & {conventional}\\
    \hline
    \textbf{Inception 5a} & {Winograd $F(2\times2, 3\times3)$}  & {-} \\
    \hline
    \textbf{Inception 5b} & {Winograd $F(2\times2, 3\times3)$}  & {-} \\
    \hline
  \end{tabular}
\end{table}
\quad We start with the conventional 6-loop unoptimized convolution as our baseline. In optimization, we use Algorithm \ref{algo_exp} to iteratively optimize the sub-branch and apply fast algorithms when possible. We implement Winograd $F(4\times4, 3\times3)$ in earlier modules and Winograd $F(2\times2, 3\times3)$ for later modules because although theoretically the former Winograd setup gives more performance gain, but in later modules, the input feature map sizes become so small that using $F(4\times4, 3\times3)$ results in lots of sampling of padded zeros, impairing both the accuracy and performance. We adopt FFT-based for \textbf{inception 3b} and \textbf{inception 4a} because the $5 \times 5$ branch becomes critical path after $3 \times 3$ branch is well optimized. CONVs in \textbf{inception 3c} and \textbf{inception 4e} have stride 2, which the fast methods don't support, so we optimize them with conventional methods \cite{zhang2017high}.\\
\indent Our implementation use 16-bit fixed point for both weight and feature map, and the operating frequency is 200 MHz. The resource consumption and simulation performance results are shown in Table \ref{resource} and Table \ref{with-fpga}.\\
\begin{table}[t]
\caption{Inception V2 resource consumption}\label{resource}
  \centering
  \begin{tabular}{|c|c|c|c|c|}
    \hline
    \textbf{} & \textbf{BRAM} & \textbf{DSP}  & \textbf{FF} & \textbf{LUT} \\
    \hline
    \textbf{Our work} & 3067 &  2041 &  539422 & 938159 \\
    \hline
    & 71\% & 32\% & 23\% & 79\% \\
    \hline
  \end{tabular}
\vspace*{-0.1in}
\end{table}
\quad We implement our design with Vivado HLS 2017.1 and find that our implementation tends to use more LUTs. This situation is due to the following reasons. First, both FFT and Winograd transformation comsume LUTs because multiplications are reduced to either additions or constant multiplications, which is implemented using LUTs. Second, the control logic in the Inception engine IP is more complicated compared to conventional convolution implementation, thus taking up more LUTs.

\subsection{Evaluation}
\quad We first compare our implementation on FPGA with GPU result. We use a cutting-edge Pascal-based NVidia GTX 1080 GPU, which has a 8.9 TFLOPS peak performance. The GPU implmentation is on Torch, with CUDA 8.0. We also compare our work with the results reported in Zhang's work (FPGA2017) \cite{zhang2017frequency}, and DiCecco's work (FPT2016) \cite{dicecco2016caffeinated} which are works that evaluated GoogLeNet (Inception V1). These two works also implement fast convolution algorithms. FPGA2017 implements Overlap-Add FFT convolver on a CPU + FPGA system, and FPT2016 implements Winograd convolution on a Xilinx Virtex7 board. Our implementation is in fact Inception V2, which is the original Inception added with batch normalization layer after each CONV layer, thus has slightly more computations. The result is shown in Table \ref{with-fpga}. We use inference latency as the evaluation metric, since facial recognition/verification is a latency critical task. For works that don't report latency, we calculate single image latency by dividing the entire network computation operations with the reported GOPS. Our result shows that, compared with GPU, we achieve 3.75x latency improvement. For FPGA works, we achieve superior results, with 3.53x and 8.11x latency speed up compared to the FPGA2017 and FPT2016, respectively. We also achieve 4.68x better energy efficiency compared to FPGA2017.

\begin{table}[t]
\caption{Inception V2 performance comparison with previous implementations}\label{with-fpga}
  \begin{adjustwidth}{-0.5cm}{}
  \small
  \begin{tabular}{|p{1.6cm}|p{1.4cm}|p{1.6cm}|p{1.4cm}|p{1.6cm}|}
    \hline
    \textbf{} & \textbf{latency per image}  & \textbf{latency speedup} & \textbf{power} & \textbf{energy efficiency}\\
    \hline
    \textbf{GPU} & 89.0 ms & 2.16X & 109.1 W & 9.780 J/pic/s\\
    \hline
    \textbf{FPT16} \cite{dicecco2016caffeinated} & 192.3 ms & 1X (baseline) & N/A & N/A\\
    \hline
    \textbf{FPGA17} \cite{zhang2017frequency} & 83.6 ms & 2.30X & 13.2 W & 1.175 J/pic/s\\
    \hline
    \textbf{Our work} & 23.7 ms & 8.11X & 10.6 W & 0.251 J/pic/s\\
    \hline
  \end{tabular}
  \end{adjustwidth}
  \vspace*{-0.15in}
\end{table}

% \begin{table}[h]
% \caption{Inception V2 latency comparison with FPGA}\label{with-fpga}
%   \centering
%   \begin{tabular}{|p{2.5cm}|c|c|c|}
%     \hline
%     \textbf{} & \textbf{Our work}  & \textbf{FPGA2017 \cite{zhang2017frequency}} & \textbf{FPT2016 \cite{dicecco2016caffeinated}}\\
%     \hline
%     \textbf{single-image performance} & 117.4 GOPS/s &  96.6 GOPS/s & 50 GOPS/s\\
%     \hline
%     \textbf{power consumption} & 10.6 W & 13.18 W & N/A\\
%     \hline
%     \textbf{power efficiency} & 11.08 GOPS/W & 7.32 GOPS/s/W & N/A\\
%     \hline
%   \end{tabular}
%   \vspace*{-0.15in}
% \end{table}

\section{Conclusions}\label{sec_conclude}
\quad In this paper, we explore different fast convolution algorithms including Winograd's minimum filter algorithm and FFT-based algorithm, and find the best strategy to apply them on different types of convolutions. We implement a configurable IP-based end-to-end CNN accelerator targeting FaceNet (Inception V2) using C-based HLS. Our solution surpasses both NVIDIA GTX 1080 GPU and previous FPGA results. We envision that such face recognition system can be paired with multiple low-power video capture systems, with the FPGA deployed in a central server and close to database, for fast real-time multi-face recognition and verification, to satisfy the need for security, border control, and other related applications. 
\begin{acks}
\quad This work is supported by IBM-Illinois Center for Cognitive Computing Systems Research (C$^3$SR), a research collaboration as part of the IBM AI Horizons Network. We also thank Kyle Rupnow of Inspirit IoT Inc. for helpful discussions.
\end{acks}

\bibliographystyle{unsrt}
\bibliographystyle{ACM-Reference-Format}
\bibliography{main_bibliography}\newpage
\end{document}